%% file: main.tex
\documentclass[runningheads]{llncs}
\usepackage[T1]{fontenc}
\usepackage{graphicx}
\usepackage{amsmath}
\usepackage{booktabs}
\usepackage[flushleft]{threeparttable}
\usepackage{xcolor}
\usepackage{cite}

\begin{document}
\title{Exploring the Meaningfulness of Nearest Neighbor Search in High-Dimensional Space}
%
%
\author{Zhonghan Chen\inst{1}, Ruiyuan Zhang\inst{1}, Xi Zhao\inst{1}, \\ Xiaojun Cheng\inst{2}, Xiaofang Zhou\inst{1}}
\authorrunning{Chen et al.}
%
\institute{Hong Kong University of Science and Technology, Hong Kong SAR \and China Unicom (Hong Kong) Operations Ltd, Hong Kong SAR\\
\email{\{zchenhj, xzhaoca\}@cse.ust.hk}\\
\email{chengxj31@chinaunicom.cn}\\
\email{\{zry, zxf\}@ust.hk}}
\maketitle              
\begin{abstract}
\input{tex_files/abstract}
\end{abstract}

\section{Introduction}
\input{tex_files/introduction_revised}

\section{Related Work}
\input{tex_files/related_work}
\section{Definition and Preliminaries}
\input{tex_files/definition}

\section{What is Meaningful Nearest Neighbor Search ?}
\input{tex_files/meaingful_revise}




\section{Experiment}
\input{tex_files/experiment_setting}

\input{tex_files/impact_metrics}

\section{Conclusion}
\input{tex_files/conclusion}

%

\bibliographystyle{splncs04}
\bibliography{myRef}

\end{document}

%% file: tex_files/abstract.tex

Dense high dimensional vectors are becoming increasingly vital in fields such as computer vision, machine learning, and large language models (LLMs), serving as standard representations for multimodal data. Now the dimensionality of these vector can exceed several thousands easily. Despite the nearest neighbor search (NNS) over these dense high dimensional vectors have been widely used for retrieval augmented generation (RAG) and many other applications, the effectiveness of NNS in such a high-dimensional space remains uncertain, given the possible challenge caused by the "curse of dimensionality." To address above question, in this paper, we conduct extensive NNS studies with different distance functions, such as $\mathcal{L}_1$-distance, $\mathcal{L}_2$-distance and angular-distance, across diverse embedding datasets, of varied types, dimensionality and modality. Our aim is to investigate factors influencing the meaningfulness of NNS. Our experiments reveal that high-dimensional text embeddings exhibit increased resilience as dimensionality rises to higher levels when compared to random vectors. This resilience suggests that text embeddings are less affected to the "curse of dimensionality," resulting in more meaningful NNS outcomes for practical use. Additionally, the choice of distance function has minimal impact on the relevance of NNS. Our study shows the effectiveness of the embedding-based data representation method and can offer opportunity for further optimization of dense vector-related applications.

\keywords{Nearest Neighbor Search \and High-Dimensional Vector \and The Curse of Dimensionality \and Embedding Model.}

%% file: tex_files/introduction_revised.tex
Nowadays, as the modalities of data become more diverse, such as text, image, and audio, it is of great significance to adopt a unified representation for these unstructured data to support various applications. Due to the success of various embedding models \cite{nvembed,gte-large,gte-qwen,croissantllm,sentence-bert,sentence-bert}, high-dimensional vector becomes a suitable solution. 
Subsequently, these vectors are effectively managed by the database to support numerous applications across domains such as including computer vision \cite{sift,alexnet}, machine learning \cite{word2vec,doc2vec}, data mining \cite{datamining1,datamining2} and retrieval augmented generation (RAG) \cite{rag1,rag2}.
As a result, the nearest neighbor search (NNS), which identifies the closest vectors from a dataset based on their distance from a query vector, has become a fundamental component of these applications.  
Despite the wide range of successful applications of nearest neighbor search in various fields, doubts remain about the meaningfulness of the algorithm, particularly in high-dimensional space. With the rise large language models, numerous embedding models have been developed that generate embeddings of dimensionality from the level of $1,000$ \cite{nvembed,gte-qwen} to even $10,000$ \cite{davinci}. 
However, the meaningfulness of performing NNS in such a high-dimensional space remains unproven as it may suffer from ``curse of dimensionality'' \cite{cod}. As the dimensionality becomes high enough, the distance between any two points tends to converge, making it difficult to distinguish between different points, solely based on distance. In fact, a proven effective and meaningful NNS is critical to various applications, such as building RAG systems for sectors like telecommunications, where accurate recall of information through NNS using dense vectors is essential \cite{tele-rag}.

To demonstrate the meaningfulness of NNS, relative contrast (RC) \cite{rc-paper} and local intrinsic dimensionality (LID)\cite{lid-paper,lid-estimate} are commonly employed \cite{ann-algorithm-related,measure-compare} to measure to what extent to which a dataset is affected by the ``curse of dimensionality''. 
RC and LID are computed based on the distance distribution of the dataset. A larger RC or a smaller LID indicates that the distance distribution deviates more from that of a random dataset, suggesting that the dataset is affected less by the ``curse of dimensionality''.
However, these studies primarily focus on the impact of RC and LID over the approximate NNS rather than the meaningfulness of the NNS itself. Moreover, they do not take the origin of the vector data into consideration. As a consequence, it is infeasible for them to demonstrate how closely the query item and its nearest neighbors correspond in original spaces, such as the text or image space. 

To address the concerns aforementioned, we conduct a comprehensive study that spans from embedding models to vector dataset, investigating factors that influence meaningfulness of NNS. Our experiments involve a total of six real-world text datasets and two image datasets. We use two distinct embedding models for generating text embeddings and the state-of-the-art CLIP model for image embeddings. 
We compute the distance distribution within each dataset by sampling a small number of query points and analyzing the RC based on distance distributions. By varying the data types, vector dimensionality, embedding models and distance functions used for NNS, we demonstrate that the choice of distance function is not a major factor affecting the meaningfulness of NNS, which neither significantly improves or degrades the meaningfulness of NNS. To further explore the effect of the ``curse of dimensionality'', we also examine the RC of the random vectors and text embeddings with varied dimensionality. The results show that, as the dimensionality increases, RC of random dataset converges to $1$ rapidly, indicating they are more liable to be affected by the ``curse of dimensionality''. In contrast, the meaningfulness of text embeddings, generated from real-world text data by embedding models, fluctuates as dimensionality increases but consistently maintains a meaningful NNS, even in high dimensional space. 
The main contributions of this paper are summarized as follows:
\begin{enumerate}
    \item We propose a comprehensive study on factors that influence the meaningfulness of NNS with high dimensional vectors. Using relative contrast, we aim to unveil the distance distributions within datasets. Our investigation examines the correlation between relative contrast and the meaningfulness of nearest neighbor search, providing insight into the extent to which the dataset is impacted by the \textit{the curse of dimensionality}.
    \item Furthermore, we conduct thorough experiments to assess how changes in dimensionality affect the meaningfulness of NNS. Our results indicate that random vectors are highly sensitive to changes in dimensionality, whereas text embeddings exhibit significant resilience, especially at higher dimensions. This indicates the effectiveness of the embedding-based data representation in the NNS applications.
    \item We carry out extensive experiments to evaluate how the choice of distance functions impacts the meaningfulness of NNS. Our findings suggest that the distance function has marginal influence on the significance of NNS outcomes. 
\end{enumerate}

For the rest of the paper, it is organized as following. Section \ref{related_work} provides an overview on previous works studying the meaningfulness of NNS. And in Section \ref{def_pre}, we provide precise mathematical definition to the relative contrast, local intrinsic dimensionality, and nearest neighbor search. In Section \ref{what_is_meaningful}, we provide meticulous discussion on what is a meaningful NNS. Then, starting from Section \ref{experiment_setting}, we discuss experiment settings of the study. And starting from Section \ref{experiment_result}, we demonstrate the experimental results and performed analysis. Lastly, in Section \ref{conclusion}, we end the paper with a brief summary of discussion.

%% file: tex_files/related_work.tex
\label{related_work}

To quantitatively measure the meaningfulness of the nearest neighbor search, relative contrast(RC) \cite{rc-paper} and local intrinsic dimensionality(LID) \cite{lid-paper}, are proposed based on the distance distribution.
given a dataset and a query points, RC depicts the ratio between the mean distance of a point in the dataset to the query and the minimal distance of a point to the query. So, when the dataset suffers from the the ``curse of dimensionality'', any a point has a similar distance to the query and RC will be close to 1. So, RC can reflect to what extent the dataset suffers from the the ``curse of dimensionality''. On the contrary, LID indicates the change ratio of the distance distribution. 
In \cite{rc-paper}, the authors illustrate the effect of RC and LID on the difficulty of ANNS and demonstrate their correlation. However, they focus little on the dataset itself and do not analyze what incurs the difference of RC and LID in the datasets.

Li et, al. \cite{ann-algorithm-related} conducted an experimental survey on ANN search with high-dimensional data, where $20$ high dimensional vector datasets are quantitatively analysed and experimented with $19$ prestigious algorithms for NNS, where RC and LID are used to evaluate the difficulty of various datasets. Despite the interaction with RC and LID, the dominant theme of the work is on benchmarking ANNS algorithms and there is no effort into the high-dimensional data itself at all.
Aumüller et, al. \cite{measure-compare} proposes a study on the influence of the LID and RC to the performance of ANN search with several traditional datasets for the evaluation of ANNS. 
Specifically, the study investigated how different distribution of LID, RC impact the performance of ANNS search algorithms. Regarding this, it showed that LID is a better predictor on performance than RC only when query workloads is light; and there does not exist a single score can predict the difference in performance. Similar as the previous one, this study focuses more on the performance of the ANNS. Also, \cite{nns-theory} investigates the characteristics of high-dimensional vectors theoretically by considering the distance distribution of random vectors. However, such an approach is hard for analyzing the real-world datasets due to their complex distributions.

%% file: tex_files/definition.tex
\label{def_pre}
\begin{definition}[Relative Contrast\cite{rc-paper}]
    Suppose \( D_{min}^q = \min_{i=1,\ldots,n} D(x_i, q) \) is the distance to the nearest database sample, and \( D_{mean}^q = {E}_x[D(x, q)] \) is the expected distance of a random database sample from the query \( q \). We define the relative contrast for the data set \( X \) for a query \( q \) as: \( C_r^q = \frac{D_{mean}^q}{D_{min}^q} \). Then, taking expectations with respect to queries, the relative contrast for the dataset $X$ is defined as:
    $$C_r = \frac{E_q[D_{mean}^q]}{E_q[D_{min}^q]} = \frac{D_{mean}}{D_{min}}$$
\end{definition}

\begin{definition}[\textbf{Local Intrinsic Dimensionality}\cite{lid-paper}]
    Let \( X \) be an absolutely continuous random distance variable. For any distance threshold \( x \) such that \( F_X(x) > 0 \), the local continuous intrinsic dimensionality of \( X \) at \( x \) is given by
\[
\text{LID}_X(x) 
= \lim_{\epsilon \to 0^+} \frac{\ln F_X((1 + \epsilon)x) - \ln F_X(x)}{\ln(1 + \epsilon)},
\]
wherever the limit exists. And the closed-form expression of LID is as following:

Let \( X \) be an absolutely continuous random distance variable. If \( F_X \) is both positive and differentiable at \( x \), then
\[
\text{LID}_X(x) = \frac{x f_X(x)}{F_X(x)}.
\]
\end{definition}






\begin{definition}[\textbf{$k$ Nearest Neighbor Search}]
Given a query point \( q \), a positive integer \( k \), and a distance metric \( d: R^d \times R^d \to R \), let \( o_i^* \) be the $i$-th exact nearest neighbor of \( q \) in \( D \). A $k$-nearest neighbor query returns a sequence of $k$ points \( \langle o_1, o_2, \ldots, o_k \rangle \) such that for each \( o_i \), we have \( d(q, o_i) \leq d(q, o_i^*) \), \( i \in [1, k] \).
\end{definition}






%% file: tex_files/meaingful_revise.tex
\label{what_is_meaningful}
In this section, we discuss criteria of a meaningful NNS with two primary concerns. Firstly, we need to identify a measure that reflects the meaningfulness without actually executing NNS algorithms. Using the measure, we examine whether the dataset is meaningful for NNS. Additionally, we assess if the NNS can retrieve similar objects in the original space, like text or image. To conclude, we consider the NNS to be meaningful if:
\begin{enumerate}
    \item Dataset is meaningful intrinsically, indicated by the measure.
    \item The NNS retrieve similar objects in the original space.
\end{enumerate}

\subsection{Meaningfulness in Intrinsic Dataset}
Currently, relative contrast (RC) \cite{rc-paper} and local intrinsic dimensionality (LID) \cite{lid-paper,lid-estimate} are two mainstream measures in evaluating the meaningfulness of the NNS. Despite both RC and LID tells similar information, which is the intrinsic property of a dataset, scopes of two measures are still different, which could be derived from the definition to RC and LID. Specifically, relative contrast gives more emphasis on describing the separation of distances between the mean and nearest neighbors, providing an insight regarding the distinctiveness of data points in NNS, which is also considered as the meaningfulness of NNS. In contrast, local intrinsic dimensionality focuses on the intrinsic dimensionality of data around query points, which describes the local complexity of datasets.

\begin{figure}[!t]
    \centering
    \includegraphics[width=\textwidth]{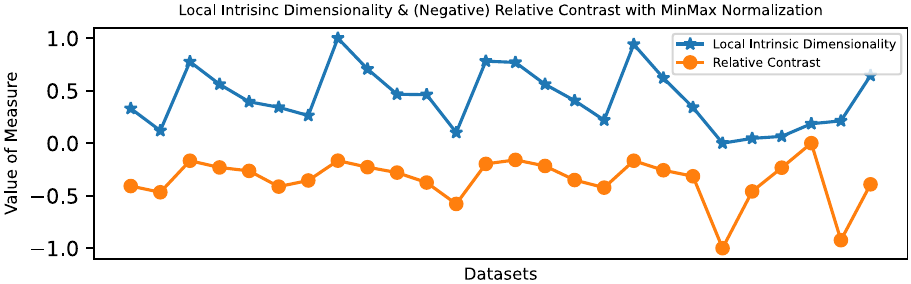}
    \vspace{-2em}
    \caption{Compare the homogeneity of RC and LID}
    \label{fig:rc-lid}
\end{figure}

To obtain a more compressive overview regarding the relation between RC and LID, we conduct an experiment, using $25$ dataset of various modalities and various dimensionality, to evaluate the homogeneity of two different metrics. In Fig.\ref{fig:rc-lid}, it shows that the RC and LID exhibit almost identical behaviour. It means that if a dataset is considered to be meaningful by one metric, the other will give the same comment, vice versa.

Therefore, considering the emphasis of scopes of RC and LID, as well as the experimental result in Fig.\ref{fig:rc-lid}, we select the relative contrast (RC) as the measure for the meaningfulness of nearest neighbor search problem. Additionally, in this subsection, we only discuss reasons that the RC is selected as the measure, and studies on the meaningfulness of various datasets, will be discussed, from different perspectives, starting from Section \ref{experiment_result}.

\subsection{Meaningfulness in Original Space}

Another criterion, beyond the quantitative measure like RC, is that the nearest neighbor search (NNS) should be able to retrieve the most similar objects in the original space, like text or image. To verify this, we perform NNS to find those vectors that have similar distances, then use results of NNS to retrieve the corresponding data from the original space.

\begin{figure}[!t]
    \centering
    \includegraphics[width=\textwidth]{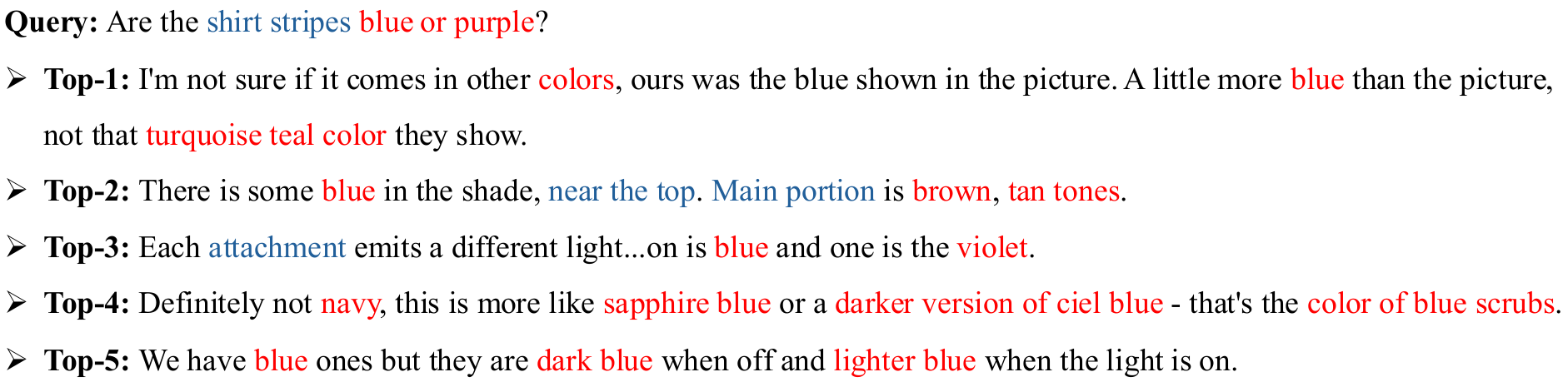}
    \caption{Example: Top-$5$ similar texts of query text retrieved by the NNS}
    \label{fig:vec2text_example}
\end{figure}

\begin{figure}[!t]
    \centering
    \includegraphics[width=\textwidth]{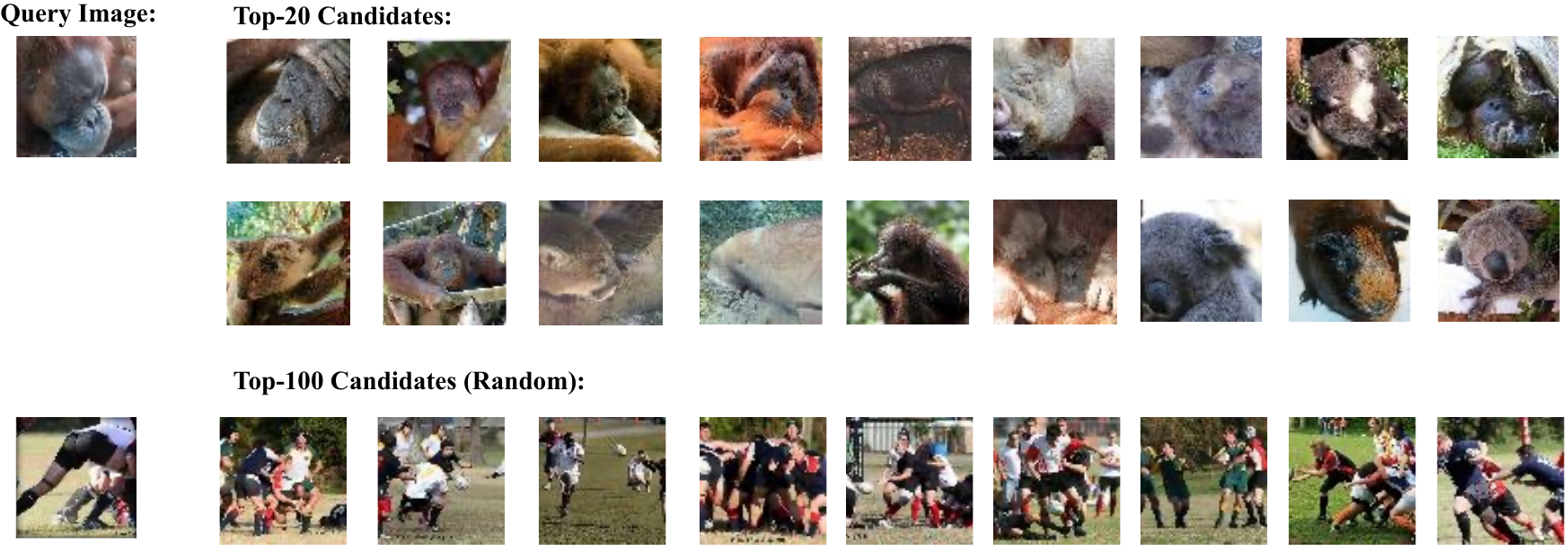}
    \caption{Example: Similar images of query image retrieved by the NNS}
    \label{fig:vec2image_example}
\end{figure}

To illustrate how these distance similar vectors appear in the original space, we present two examples of two modalities: Fig. \ref{fig:vec2text_example} and Fig. \ref{fig:vec2image_example}, where NNS effectively retrieves these similar objects within the original space.

%% file: tex_files/experiment_setting.tex
\subsection{Datasets}
\label{experiment_setting}
In this study, we include several types of different vector data in the high-dimensional space. Overall, we primarily focus on the following: synthesized random vector datasets, following Gaussian distribution; vectors extracted from image feature using traditional algorithm in computer vision (SIFT1M\cite{sift} and GIST1M\cite{gist}); text embeddings are generated with prestigious embedding models\cite{sentence-bert,miniLM} with real-world text datasets\cite{AgNews,AmazonQA,GooAQ,Yahoo,WikiSummary,Orca-Chat}. To further diversify modalities of datasets, we include two image embedding datasets\cite{imagenet,Places2}, generated by the State-of-the-Art CLIP \cite{CLIP} by OpenAI. Technical details of datasets could be found with Table \ref{tab:dataset info}.

\input{tables/dataset}
\vspace{-2em}

\subsection{Embedding Models}

In this study, we employ two prominent text embedding models: all-MiniLM-L6-V2 \cite{miniLM} and bert-base-nli-mean-tokens\cite{sentence-bert}, both from the sentence-transformer library. Details of these models are provided in the table below. 
\begin{table}[b]
    \centering
    \caption{Technical Details: Text Embedding Models}
    \begin{threeparttable}
        \begin{tabular}{|l|c|c|c|c|}
            \toprule
            \textbf{Model Name} & \textbf{Param Size} & \textbf{Max Input Token} & \textbf{Dimensionality} \\
            \midrule
            all-MiniLM-L6-V2 & 22.7M & 256 & 384 \\
            bert-base-nli-mean-tokens & 109M & 128 & 768\\
            \bottomrule
        \end{tabular}
    \end{threeparttable}
    
    \vspace{0.5em}
    \label{tab:text_model}
    \vspace{-1.5em}
\end{table}


\subsection{Evaluation Metric \& Factors}

\textbf{\underline{Metric:}}
In this study, the meaningfulness of the nearest neighbor search is evaluated by the relative contrast (RC). \\
\textbf{\underline{Factors:}}
We firstly explore if different distance functions would influence the meaningfulness and the result of the NNS. This investigation considers data of various types and dimensionalities, including random vectors, image feature vectors, image embeddings, and text embeddings. Next, we examine how changes in dimensionality influences the meaningfulness of the NNS, where we specifically focus on the random vector and text embeddings. In short, two factors of the evaluation are: distance function and the change in dimensionality.

%% file: tables/dataset.tex
\begin{table}[ht]
    \small
    \centering
    \caption{Information of Raw Datasets}
    \begin{tabular}{|l|c|c|c|r|}
        \toprule
        \textbf{Dataset} & \textbf{Dim.} & \textbf{Card.} & \textbf{Description}  \\
        \midrule
        RANDOM & $16-4096_{1}$ & $1$M & Random following Gaussian Distribution \\
        SIFT1M \cite{sift} & $128$ & $1$M & SIFT descriptor on Image\\
        GIST1M \cite{gist} & $960$ & $1$M & GIST descriptor on Image \\
        ImageNet-Tiny \cite{imagenet} & $512$ & $100,000$ & $200$ classes of $64\times64$ images of various classes \\ 
        Places2 \cite{Places2} & $512$ & $1,803,460$ & A large-scale database for scene understanding \\ 
        AmazonQA \cite{AmazonQA} & $384-12288_{2}$ & $1,095,290$ & Amazon product review \\ 
        WikiSummary \cite{WikiSummary} & $384-12288$ & $5,315,384$ & Wikipedia items containing name and abstract \\                   
        GooAQ \cite{GooAQ} & $384-12288$ & $3,012,496$ & Google Auto Suggest in Q\&A format\\ 
        AgNews \cite{AgNews} & $384-12288$ & $1,157,745$ & News article corpus with title and abstract \\ 
        Yahoo \cite{Yahoo} & $384-12288$ & $1,198,260$ & Pairs of title and answer from Yahoo \\ 
        OrcaChat \cite{Orca-Chat}& $384-12288$ & $862,046$ & Orca Chat Dataset (Q\&A pairs) for LLM SFT \\ 
        \bottomrule
    \end{tabular}
    \begin{tablenotes}
        \small
        \item \textbf{Note:}
        \item \begin{enumerate}
            \item The dimensionality of random datasets are: $16$, $32$, $64$, $128$, $384$, $512$, $768$, $1024$, $2048$, $3584$, $4096$, most of which are common values for various embedding models, according to MTEB Benchmark\cite{mteb-benchamrk}.
            \item The dimensionality of text embedding datasets are: $384$, $512$, $768$, $1024$, $2048$, $3584$, $8192$, $12288$.
        \end{enumerate}
    \end{tablenotes}
    
    \label{tab:dataset info}
\end{table}

%% file: tex_files/impact_metrics.tex
\label{experiment_result}
\subsection{Distance Function}
\begin{figure}[!t]
    \centering
    \includegraphics[width=\textwidth]{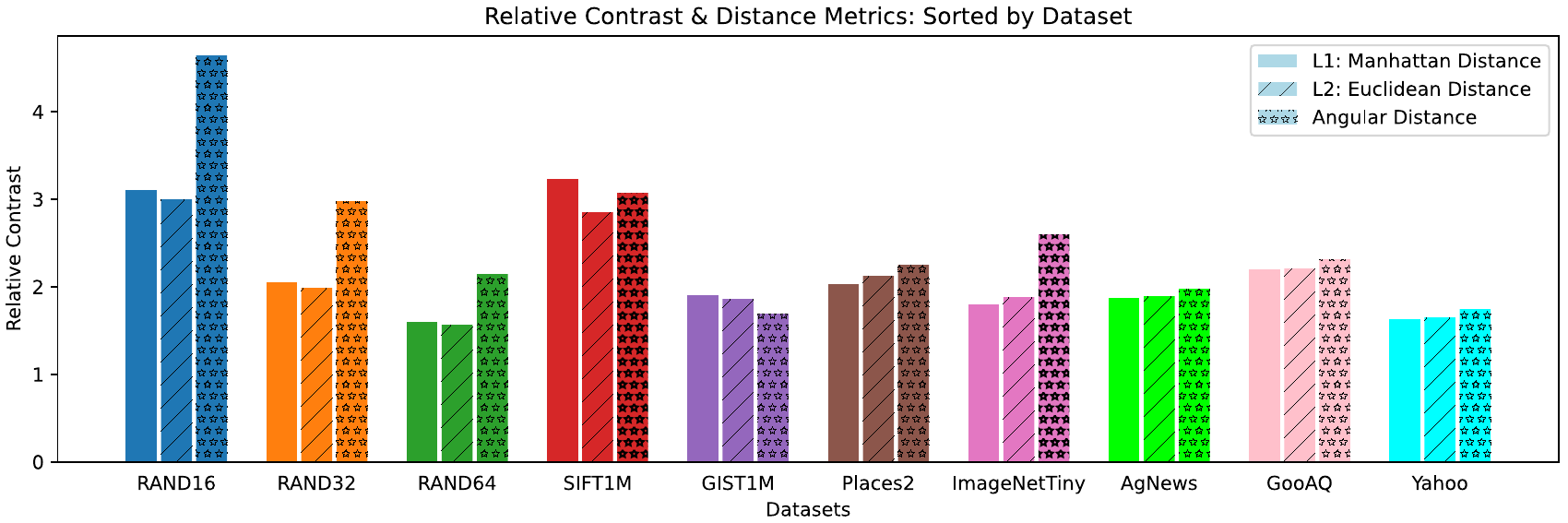}
    \includegraphics[width=\textwidth]{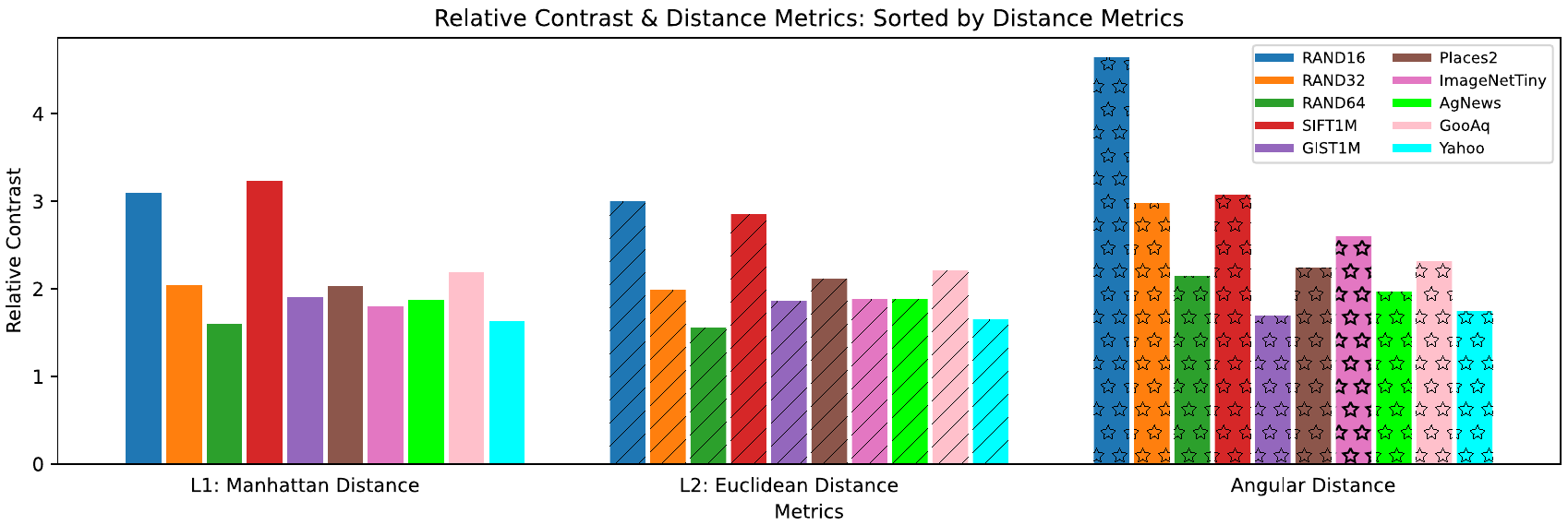}
    \caption{Explore the impact of distance function on the Relative Contrast (Upper: sort by dataset || Lower: sort by function)}
    \label{fig:metric}
\end{figure}

Firstly, with Fig. \ref{fig:metric}, it is obvious that distinct distance functions indeed result in different values of the relative contrast for each dataset, and thus different performance in NNS. Despite different distance functions give rise to distinct value of RC for each dataset, it does not actually have any significant impact on the overall ranking of each dataset, given a distance function. Specifically, it means that if the RC of dataset A is higher than the RC value of dataset B, under one of the distance functions, the relation remains mostly the same under other distance measure as well.

From the perspective of the datasets, we notice that the random datasets are more vulnerable to the change in the distance function, especially when the dimensionality is low. On the contrary, other types of vectors are relatively more stable when different distance functions are used. From the perspective of the value of RC itself among these datasets, it could be easily observed that RC values, under $\mathcal{L}_1$ and $\mathcal{L}_2$ distance, are more similar than the angular distance. And the phenomenon could be explained by the nature of $\mathcal{L}_p$ norm, and how the distance is computed. 

Moreover, according to Fig. \ref{fig:metric}, we may derive that changing the distance function will not significantly impact the meaningfulness of the dataset in NNS. Specifically, it means that changing different distance function will not necessarily improve or degrade the meaningfulness of the dataset in nearest neighbor search.

In short, we know that distance function is not a major concern that influences the meaningfulness of the nearest neighbor search problem. 

\subsection{Dimensionality of High-Dimensional Vector}

Dimensionality is one of the most crucial property of vector data, particularly in the high-dimensional space. With the great success of large language models and learning models, the dimensionality of vector / embedding has increased to a completely unprecedented level. For instance, according to the MTEB benchmark \cite{mteb-benchamrk}, dozens of embedding models generates embeddings with dimensionality greater than $1,000$ \cite{nvembed,gte-qwen,gte-large,bigembed_model,bigembed_model2}, where some popular dimensionalities are: $1024$, $2048$, $3584$, $4096$, and there are even few models generating vectors of dimensionality over $10,000$ \cite{davinci}. There is evidence suggesting that, in the context of LLM, vector with higher dimnsionality are better at capturing complex knowledge from text compared to lower-dimensional vectors. However, does a higher dimensionality always lead to better performance and a more meaningful NNS ? And to what extent might it suffer from the ''curse of dimensionality'' ?

In this section, we focus on how the increase in dimensionality would impact the meaningfulness of the NNS for two types of vectors: synthesized random vector and text embedding, in terms of the relative contrast (RC).

Additionally, we have not included a discussion on image embeddings in this study. The reason is that text remains the primary modality in current LLM-RAG systems, and image data is yet a major focus at the moment. Moreover, adjusting the dimensionality of an image embedding model is particularly complex, which requires significant efforts in retraining the model and conducting extensive evaluation.

\subsubsection{Synthesized Random Vector:}

\begin{figure}[!t]
    \centering
    \includegraphics[width=\textwidth]{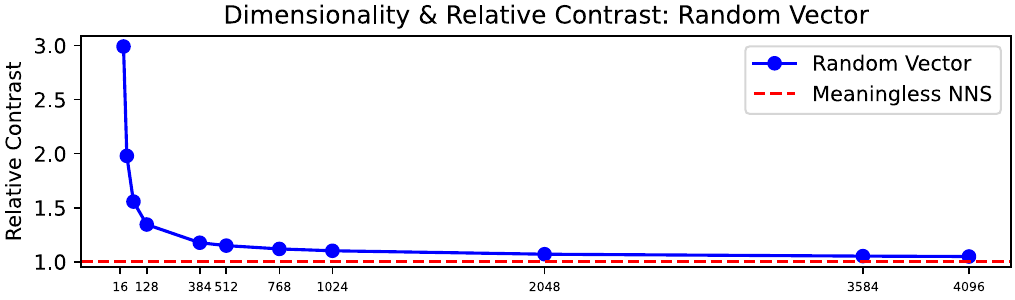}
    \caption{Explore the impact of dimensionality on high-dimensional random vector}
    \label{fig:dim_random}
\end{figure}

We start the section with the simplest type of vector, which is the randomly synthesized vectors. And the result of the experiment is shown in Fig. \ref{fig:dim_random}. Specifically, in this set of the experiment, random vector of a wide range of dimensionalities are involved, including: $16$, $32$, $64$, $128$, $384$, $512$, $768$, $1024$, $2048$, $3584$, and $4096$. These values of the dimensionality are commonly used in various embedding models, according to the MTEB benchmark\cite{mteb-benchamrk}.

Specifically, with Fig. \ref{fig:dim_random}, we notice that the value of relative contrast is almost $3$ when the dimensionality is $16$, which indicates an extremely meaningful nearest neighbor search. As the dimensionality of the vector is further increased to $128$, it is obvious that the curve becomes steep, where a sharp decrease of the RC occurs. Despite the dimensionality of random vectors is low (like $64$ and $128$), which is incomparable with those popular text embeddings at the same dimensionality, the meaningfulness of the NNS still does not degrades significantly. As we further increase the dimensionality to $384$ and eventually $4096$, it is obvious that the curve quickly converges. And the value of the curve is extremely close to $1$, which indicates an almost meaningless NNS.

In short, with this set of experiment, we know that the meaningfulness of the NNS of random vector degrades significantly as dimensionality increases, and the NNS quickly converges to mostly meaningless when the dimensionality is around $512$ and $768$.

\subsubsection{High-Dimensional Text Embedding}

\begin{figure}[!t]
    \centering
    \includegraphics[width=\textwidth]{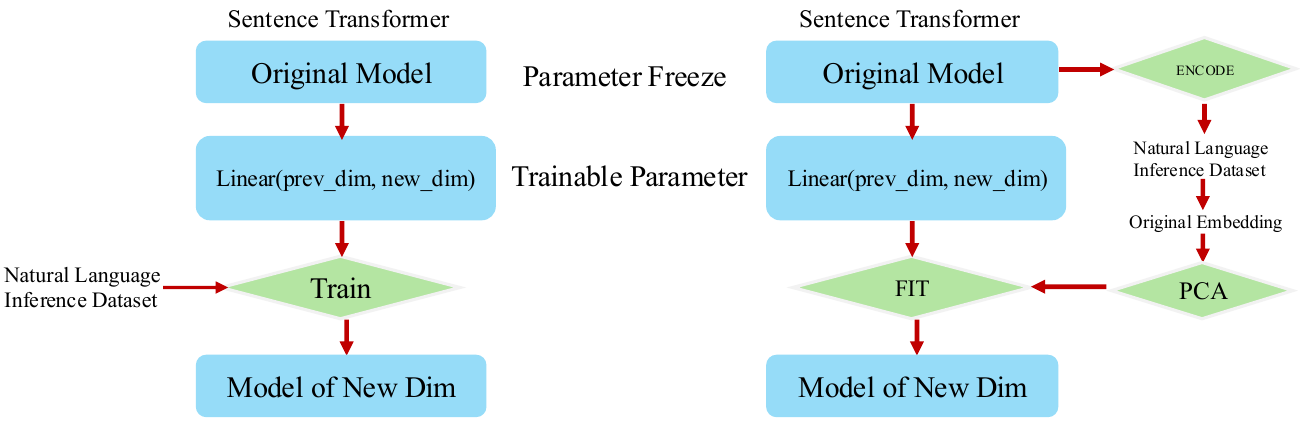}
    \caption{Workflow of Increasing the Dimensionality of Embedding Model}
    \label{fig:workflow}
\end{figure}

In the previous section, we observe that the randomly synthesized vector becomes meaningless in NNS as dimensionality increases and the value of the relative contrast rapidly converges to $1$ in low dimensional space. Now, we are wondering, how the RC of text embeddings varies as the dimensionality increases. 

To answer the question above, we investigate into the behavior of the text embeddings, in terms of relative contrast. Technically, two prestigious models in sentence-transformer library are involved in experiments: all-MiniLM-L6-v2 \cite{miniLM} and bert-base-nli-mean-tokens \cite{sentence-bert}. To customize the dimensionality of the same embedding model, a dense layer is attached to the end of the embedding model, mapping the vector from the original space to the space with desired dimensionality. Then, all trainable parameters of the original model are frozen, and popular natural language inference (NLI) datasets \cite{multinli,snli} are used to train the newly added dense layer with identical training parameters.

Another approach to derive an embedding model with customized dimensionality is similar to the previous one, where a dense layer is attached to the end of the the original model as well. The difference is that, the weight of the dense layer is computed by performing the principal component analysis (PCA) on the embedding of the original vector, instead of the learning approach aforementioned. Such method effectively achieves the goal while less time and computational resources are required. However, the method is a lossy transformation due to the nature of PCA, comparing with the training method, which is still a process of continuous optimization. Considering this, only the first approach (training) is utilized to customize the dimensionality of the embedding model to guarantee the effectiveness and fairness.

\begin{figure}[!t]
    \centering
    \includegraphics[width=\textwidth]{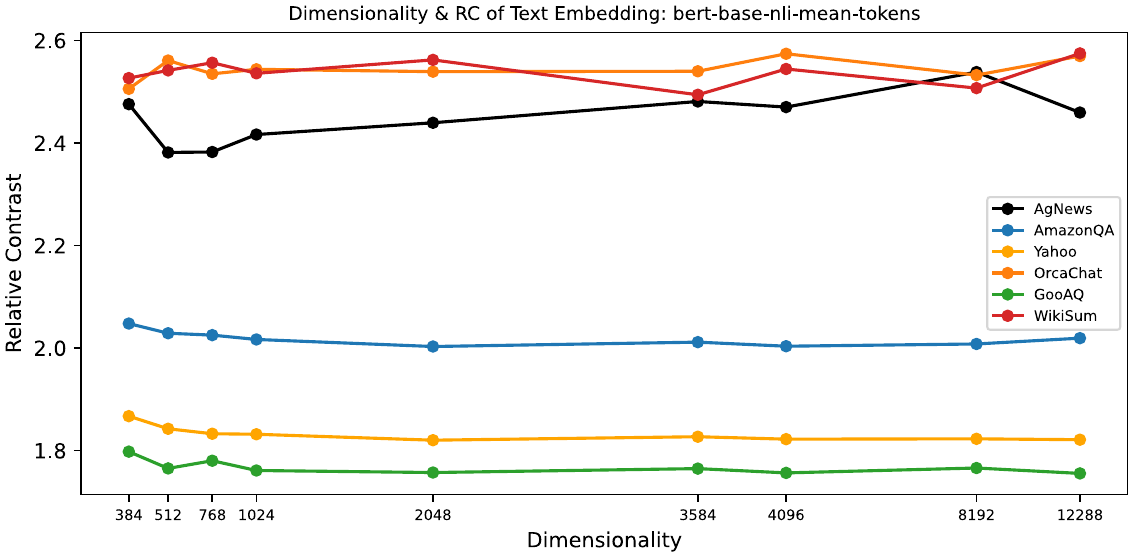}
    \caption{Impact of dimensionality on high-dimensional text embedding with bert-base-nli-mean-tokens model}
    \label{fig:dim_text2}
\end{figure}

\begin{figure}[!t]
    \centering
    \includegraphics[width=\textwidth]{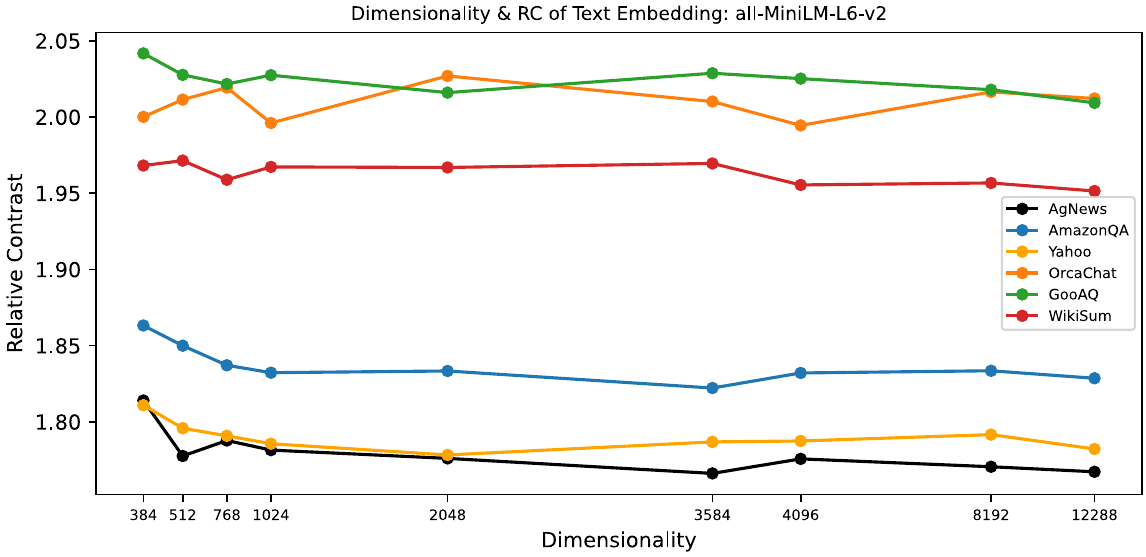}
    \caption{Impact of dimensionality on high-dimensional text embedding with all-MiniLM-L6-v2}
    \label{fig:dim_text1}
\end{figure}

Now, let us turn to the result of the experiment. In this part of the experiment, we firstly use the bert-base-nli-mean-tokens \cite{sentence-bert} model, which is a commonly used embedding model based on the prestigious ''Bidirectional Encoder Representations from Transformers'' (BERT) \cite{bert-paper} model. With Fig. \ref{fig:dim_text1}, we can easily observe that, for each dataset, as the dimensionality of embeddings increases, there is no evidence showing that the value of relative contrast will monotonically decrease. Moreover, during some certain interval of dimensionality, the relative contrast of the dataset increases as dimensionality increases. Overall, the RC fluctuates as the dimensionality increases; and from the perspective of the absolute value, RC values are consistently maintained at a high value (from $1.75$ to $2.05$), which indicates a meaningful NNS.

To enhance the robustness of the experiment, we conduct another experiment using the all-MiniLM-L6-v2 \cite{miniLM} as the base model. And this model is one of the smallest model in the area, containing only $22.7$M parameters while the BERT model contains $109$M parameters. As for the result of experiments, curves of relative contrast exhibit similar tendency as the previous one, where we further demonstrate and prove that increasing the dimensionality of the text embedding will not result in significantly negative impact on the performance of the NNS, in terms of the relative contrast. 

In short, based on the experiment in the section, we find that the increasing the dimensionality of the text embedding into space of higher dimensionality will not necessarily degrade the performance and meaningfulness of the NNS. And the value of relative contrast is consistently maintained at a descent level, which directly indicates a meaningful NNS.

%% file: tex_files/conclusion.tex
\label{conclusion}
In this paper, we provide an exploration on the factor that affect the meaningfulness of high-dimensional vectors in the NNS problem. Specifically, with carefully tailored real-world datasets in the high-dimensional space, we focus on the impact brought by the choice of distance function and dimensionality. Our experimental results suggest that distance function is not a major concern that impacts the meaningfulness of the nearest neighbor search. More importantly, our experiments indicate that the increment of dimensionality would significantly degrade the meaningfulness of NNS for random vectors. However, the increment of dimensionality on text embeddings has minor effects over the meaningfulness of the NNS. Moreover, even when there is a tendency of decrease, the text embedding is still able to maintain a meaningful nearest neighbor search.